\documentclass[12pt]{article}
\usepackage{amsmath}
\usepackage{graphicx}
\usepackage{natbib}
\usepackage{url} 

\usepackage[utf8]{inputenc} 
\usepackage[T1]{fontenc}    
\usepackage[hidelinks]{hyperref}       
\usepackage{booktabs, makecell}       
\usepackage{amsfonts, amsthm}       
\usepackage{nicefrac}       
\usepackage{microtype}      
\usepackage{comment}
\usepackage{subcaption}
\usepackage{xcolor}
\usepackage{amsmath}
\usepackage{mathtools}
\usepackage[title]{appendix}
\usepackage{bbm}
\usepackage{multirow}
\usepackage{enumerate}
\usepackage{float}
\usepackage{amssymb}
\usepackage{siunitx}
\usepackage{dcolumn}
\newcolumntype{d}[1]{D{.}{.}{#1}}
\usepackage{enumitem}
\setlist[itemize]{leftmargin=1.2cm}
\usepackage{algorithm2e}
\usepackage{algorithmic}
\usepackage{caption}

\newtheorem{proposition}{Proposition}

\newtheorem{definition}{Definition}

\usepackage{booktabs,siunitx}
\newcommand{\deepmomR}{DeepMoM}
\newcommand{\deepmom}{\texttt{\deepmomR}}
\newcommand{\samxi}{\boldsymbol{x}_i}

\newcommand{\inputx}{\boldsymbol{x}}
\newcommand{\outcomey}{\boldsymbol{y}}
\newcommand{\radius}{\boldsymbol{r}}
\newcommand{\numclass}{c}
\newcommand{\valuea}{a}
\newcommand{\robust}{k}

\newcommand{\mom}{\operatorname{MoM}}

\newcommand{\se}{\operatorname{SE}}
\newcommand{\abe}{\operatorname{AD}}
\newcommand{\hue}{\operatorname{H}}
\newcommand{\sce}{\operatorname{SCE}}

\newcommand{\imax}{i_{\max}}
\newcommand{\stopping}{d}
\newcommand{\rate}{r}
\newcommand{\batch}{h}

\makeatletter
\newcommand *{\eqv}{\mathrel{\rlap{\raisebox{0.3ex}{$\m@th\cdot$ }}\raisebox {-0.3ex}{$\m@th\cdot$}}=}
\makeatother

\newcommand{\numneuron}{\overline{p}}

\newcommand{\inputv}{\boldsymbol{x}}
\newcommand{\target}{\boldsymbol{t}}
\newcommand{\err}{\boldsymbol{u}}

\newcommand{\func}{f}

\newcommand{\Mmin}{\operatorname{MoM}_{\operatorname{min}}}
\newcommand{\MCV}{\operatorname{MoM}_{\operatorname{CV}}}
\newcommand{\Hmin}{\operatorname{Huber}_{\operatorname{min}}}
\newcommand{\Logone}{\operatorname{Log}_{1}\operatorname{CV}}
\newcommand{\Logtwo}{\operatorname{Log}_{2}\operatorname{CV}}
\newcommand{\net}{\mathfrak{g}}
\newcommand{\Loss}{\mathfrak{L}}

\newcommand{\block}{\mathcal{B}}
\newcommand{\class}{\mathcal{C}}
\newcommand{\normal}{\mathcal{N}}
\newcommand{\numblock}{b}

\newcommand{\actif}{\mathbf{\mathfrak{f}}}
\newcommand{\prob}{P}

\newcommand{\weight}{W}
\newcommand{\weights}{\boldsymbol{W}}

\newcommand{\matri}{M}
\newcommand{\matrices}{\boldsymbol{M}}
\newcommand{\matrixspace}{\mathcal{M}}
\newcommand{\bias}{\theta}
\newcommand{\biases}{\boldsymbol{\theta}}
\newcommand{\biaspace}{\Theta}
\newcommand{\quantile}{Q}

\newcommand{\info}{\mathcal{I}}
\newcommand{\outlier}{\mathcal{O}}

\newcommand{\numsample}{n}
\newcommand{\numparameter}{p}
\newcommand{\numlayer}{l}

\newcommand{\R}{\mathbb{R}}

\newcommand{\bc}[1]{\ensuremath{\{{#1}\}}} 
\newcommand{\bcb}[1]{\ensuremath{\big\{{#1}\big\}}} 
\newcommand{\bcbb}[1]{\ensuremath{\Big\{{#1}\Big\}}} 
\newcommand{\bcbbbb}[1]{\ensuremath{\Bigg\{{#1}\Bigg\}}} 

\newcommand{\prt}[1]{\ensuremath{({#1})}} 
\newcommand{\prtb}[1]{\ensuremath{\big({#1}\big)}} 
\newcommand{\prtbbbb}[1]{\ensuremath{\Bigg({#1}\Bigg)}} 

\newcommand{\bs}[1]{\ensuremath{[{#1}]}} 
\newcommand{\bsb}[1]{\ensuremath{\big[{#1}\big]}} 
\newcommand{\bsbbbb}[1]{\ensuremath{\Bigg[{#1}\Bigg]}} 

\newcommand{\norm}[1]{\ensuremath{|\!|#1|\!|}} 

\DeclareMathOperator*{\argmin}{arg\,min} 

\newcommand{\blind}{0}

\begin{document}

\def\spacingset#1{\renewcommand{\baselinestretch}%
{#1}\small\normalsize} \spacingset{1}


\if0\blind
{
  \title{\bf \deepmomR: Robust Deep Learning With Median-of-Means}
  \author{Shih-Ting Huang \\
    Department of Mathematics,
  Ruhr University, Bochum\\
    and \\
    Johannes Lederer \\
  Department of Mathematics,
  Ruhr University, Bochum\\}
  \maketitle
} \fi

\if1\blind
{
  \bigskip
  \bigskip
  \bigskip
  \begin{center}
    {\LARGE\bf Title}
\end{center}
  \medskip
} \fi

\bigskip
\begin{abstract}
Data used in deep learning is notoriously problematic.
For example, 
data are usually combined from diverse sources,
rarely cleaned and vetted thoroughly,
and sometimes corrupted on purpose.
Intentional corruption that targets the weak spots of algorithms has been studied extensively under the label of ``adversarial attacks.''
In contrast, the arguably much more common case of corruption that reflects the limited quality of data has been studied much less.
Such ``random'' corruptions are due to measurement errors, unreliable sources, convenience sampling, and so forth.
These kinds of corruption are common in deep learning, 
because data are rarely collected according to strict protocols---in strong contrast to the formalized data collection in some parts of classical statistics.
This paper concerns such corruption.
We introduce an approach motivated by very recent insights into median-of-means and Le Cam's principle,
we show that the approach can be readily implemented,
and we demonstrate that it performs very well in practice.
In conclusion, we believe that our approach is a very promising alternative to standard parameter training based on least-squares and cross-entropy loss.
\end{abstract}

\noindent%
\vfill

\newpage
\spacingset{1.5} 

\section{Introduction}
\label{sec:Introduction}

Deep learning is regularly used in safety-critical applications.
For example,
deep learning is used in the object-recognition systems of autonomous cars,
where malfunction may lead to severe injury or death.
It has been shown that data corruption
can have dramatic effects on such critical deep-learning pipelines  \citep{akhtar2018threat,8611298, kurakin2016adversarial, wang2019direct, 10.1145/2976749.2978392, tksl, kurakin2016adversarial2}.
This insight has sparked research on robust deep learning based on, for example, adversarial training~\citep{aleks2017deep, kos2017delving, papernot2015distillation, tramr2017ensemble, salman2019provably},  sensitivity analysis~\citep{wang2018robust}, or noise correction~\citep{patrini2017making,Yi_2019_CVPR,pmlr-v97-arazo19a}.

Research on robust deep learning focuses usually on ``adversarial attacks,'' 
that is, intentional data corruptions designed to cause failures of specific pipelines. 
In contrast, the fact that data is often of poor quality much more generally has received little attention.
But low-quality data is very common,
simply because  data used for deep learning is rarely collected based on rigid experimental designs but rather amassed from whatever resources are available~\citep{8862913,friedrich2020role}.
Among the few papers that consider such corruptions are~\citep{barron2019general, belagiannis2015robust, jiang2018mentornet, wang2016studying, lederer2020risk}, who replace the standard loss functions, such as squared error and soft-max cross entropy loss functions by some Lipschitz-continuous alternatives, such as Huber loss functions.
But there is much room for improvement,
especially because the existing methods do not make efficient use of the uncorrupted samples in the data.

In this paper, we devise a novel approach to deep learning with ``randomly'' corrupted data.
The inspiration is the very recent line of research on median-of-means~\citep{lugosi_mendelson_2019a,lugosi_mendelson_2019b, lecue_lerasle_mathieu_2020} and Le Cam’s procedure~\citep{lecam_1973, cam_1986} in linear regression \citep{guillaume2017learning,Lecu2017RobustML}.
Specifically,
we establish parameter updates in a min-max fashion,
and we show that this approach indeed outmatches other approaches on simulated and real-world data.

\pagebreak
Our three main contributions are:
\begin{itemize}
    \item We introduce a robust training scheme for neural networks that incorporates the median-of-means principle through Le Cam's procedure. 
    \item We show that our approach can be implemented by  using a simple  gradient-based algorithm.  
    \item We demonstrate that our approach outperforms standard deep-learning pipeline across different levels of corruption.  
\end{itemize}

\paragraph{Outline of the paper} 
In Section~\ref{sec:real}, we state the motivation of deriving our $\deepmom$ estimator and demonstrate the advantages of it in prediction on real data.
In Section~\ref{sec:estimator}, we state the problem and give a step-by-step derivation of our $\deepmom$ estimator (Definition~\ref{def:estimator}).
In Section~\ref{sec:RelateLiterature},
we highlight similarities and differences to other approaches.
In Section~\ref{subsec:numerical}, we establish a stochastic-gradient algorithm to compute our estimator  (Algorithm~\ref{alg:mom}).
In Sections~\ref{subsec:simulation1} and~\ref{subsec:simulation2}, we demonstrate that our approach rivals or outmatches traditional training schemes based on least-squares and on cross-entropy for both corrupted and uncorrupted data.
Finally, in Section~\ref{sec:discussion}, we summarize the results of this work and conclude.


\section{Motivation}
\label{sec:real}
Before we formally derive our robust estimator, we first illustrate its potential in practice.
We consider classification of cancer types based on microRNA data obtained from the well-known TCGA Research Network~\citep{tcgadatabase}.
The specific data are a combination of seven TCGA projects as summarized in Table~\ref{table:datasummarize}. 
The data comprises $\numsample=2597$ samples with each $\numparameter=1881$  features.
Each sample corresponds to one of the seven labels (cancer types) shown in the table.
The data are normlized by the total count normalization method proposed by~\citep{10.1093/bib/bbs046}. 
The network settings are the ones described later in Section~\ref{subsec:simulation1}.

\begin{table}[ht]
\centering
	\begin{tabular}[t]{l c}
	
		\toprule
		\midrule
		
		Project ID (cancer type) & 
	    Number of samples \\
		
		\hline
	 
	    TCGA-OV    & 489     \\
	    TCGA-SARC  & 259     \\ 
	    TCGA-KIRC  & 516     \\
	    TCGA-LUAD  & 513     \\ 
	    TCGA-SKCM  & 448     \\
	    TCGA-ESCA  & 184     \\
	    TCGA-LAML  & 188     \\
		\midrule
		\bottomrule
	\end{tabular}
	\caption{summarization of the TCGA datasets for analysis} 
	\label{table:datasummarize}  
\end{table}

The goal is to predict the labels. 
We randomly partition the original data into ten subsets and then take the average over ten analyses,
where always one subset is used for validation and the remaining data for training.

We compare our robust deep-learning approach  (defined later in  Section~\ref{subsec:simulation1}; called $\MCV$ here) with a standard deep-learning approach (soft-max cross entropy; called SCE here) and 
two classical methods (logistic regression with $\ell_{1}$ and $\ell_{2}$ regularization; called $\Logone$ and $\Logtwo$ here).
The number of blocks~$\numblock$ for our method is chosen by $10$-fold cross-validation.
In addition, we also compute the $\deepmom$ estimator with the number of block $\numblock$ (defined in Section~\ref{sec:estimator}) selected by 10-fold cross validation of the whole data and denote it by $\MCV$.
Similarly,
For the logistic estimators with $\ell_{1}$ and $\ell_{2}$ regularizations, we use $10$-fold cross validation to select among $300$~tuning parameters within \begin{equation*}
    \biggl\{
    \tau 
    \>|\> 
    \tau 
    = 
    0.1 + 
    \frac{29.9}{299} \times i,
    ~~~~~ i\in\{0,\dots,299\} 
    \biggr\}\,.
\end{equation*}

\begin{table}[ht]
\centering
	\begin{tabular}[t]{l l}
	
		\toprule
		\midrule
		
		\multicolumn{2}{c}{Classification on TCGA data} \\ 
		\cline{1-2}
		
		Method & 
	    Accuracy \\
		
		\hline
	 
	    $\MCV$~~~~~~~~~~&\textbf{91.49\,\%}     \\ 
	    $\sce$ & 78.12\,\%    \\
	    $\Logone$  & 78.78\,\%     \\ 
	    $\Logtwo$ & 84.37\,\%    \\
		\midrule
		\bottomrule
	\end{tabular}
	\caption{$\MCV$ outperforms their competitors in prediction} 
	\label{table:real}  
\end{table}

The results are summarized in Table~\ref{table:real}.
These results suggest that the $\deepmom$ estimator is indeed much more accurate than its competitors.
Why is this the case?
Recent research has revealed that TCGA data is often normally distributed on some parts but far from normally distributed on other parts,
with heavy tails and other issues \citep{Torrent2020TheSO} on those latter parts of the data.
Standard methods, both traditional and deep-learning methods,
are not able to cope with such data. 
Our $\deepmom$,
on the other hand, partitions the data into blocks and takes a robust median over these blocks,
which ensures both an efficient use of the benign samples and an robust treatment of the problematic samples.
We will see in the following that these benefits apply to a wide range of data.


\section{Framework and estimator}
\label{sec:estimator}
We first introduce the statistical framework and our corresponding estimator.
We consider data $\prt{\outcomey_{1},\inputx_1},\dots,\prt{\outcomey_{\numsample}, \inputx_{\numsample}} \in \R^{\numclass} \times \R^{\numparameter}$ with $\numclass,\numparameter \in \bc{1,2,\dots}$ such that 
\begin{equation}
\label{eq:data}
    \outcomey_{i} = 
    \net_{*}\bs{\inputx_{i}}
    +\err_{i} ~~~~~ i \in \bc{1, \dots, \numsample},
\end{equation}
\noindent where $\net_{*} : \R^{\numparameter} \mapsto \R^{\numclass}$ is the unknown data-generating function, 
and $\err_{i} \in \R^{\numclass}$ are the stochastic error vectors. 
In particular,
each $\inputx_i$ is an input of the system and $\outcomey_i$ the corresponding output. 
The data is partitioned into two parts:
the first part comprises the informative samples;
the second part comprises the problematic samples (such as corrupted samples---irrespective of what the source of the corruption is).
The two parts are index by $\info\subset\{1,\dots,\numsample\}$ and  $\outlier\eqv\{1,\dots,\numsample\}\setminus\info$,
respectively.
Of course, the sets~$\info$ and~$\outlier$ are unknown in practice (otherwise, one could simply remove the problematic samples).
In brief,
we consider a standard deep-learning setup---with the only exception that we make an explicit distinction between ``good'' and ``bad'' samples.

Our general goal is then, as usual, to  approximate the data-generating function $\net_{*}$ defined in~\eqref{eq:data}.
But our specific goal is to take into account the fact that there may be problematic samples.
Our candidate functions are feed-forward neural networks $\net_{\matrices,\biases}$ of the form
\begin{equation}
\label{eq:nn}
\net_{\matrices,\biases} \bs{\inputv} 
\eqv 
\matri^{\numlayer} \actif^{\numlayer} \bsb{
            \matri^{\numlayer - 1}
            \dots \actif^{1} \bs{
                \matri^{0} \inputv + \bias^{0}
            }
        +\bias^{\numlayer-1}
        }
        +\bias^{\numlayer}
        ~~~~~\inputv \in \R^{\numparameter},
\end{equation}
\noindent indexed by the parameter spaces
\begin{equation*}
    \matrixspace
    \eqv
    \Big\{
        \matrices=(\matri^{0},\dots,\matri^{\numlayer}):\matri^{j}\in \R^{\numparameter^{j+1} \times \numparameter^{j}} \text{for all} ~~ j \in \bc{0,\dots,\numlayer}
    \Big\}
\end{equation*}
\noindent and 
\begin{equation*}
    \biaspace
    \eqv
    \Big\{
        \biases=(\bias^{0},\dots,\bias^{\numlayer}):\bias^{j}\in \R^{\numparameter^{j+1}}  \text{for all} ~~ j \in \bc{0,\dots,\numlayer}
    \Big\},
\end{equation*}
\noindent where 
$\numlayer$ is the number of layers,
$\matri^{j} \in \R^{\numparameter^{j + 1} \times \numparameter^{j}}$ are the (finite-dimensional)  weight matrices 
with $\numparameter^{\numlayer + 1}=\numclass$ and $\numparameter^{0}=\numparameter$, $\bias^{j} \in \R^{\numparameter^{j+1}}$ are bias parameters,
and $\actif^{j}:\R^{\numparameter^{j+1}}\to\R^{\numparameter^{j+1}}$ are the activation functions.
For ease of exposition,
we concentrate on ReLU activation, that is, $\prt{\actif^{j}\bs{\inputv}}_{i} \eqv \max\bc{0,\prt{\inputv}_{i}}$ for $\inputv \in \R^{\numparameter^{j+1}}$, $i \in \bc{1,\dots,\numparameter^{j+1}}$, and $j \in \bc{1, \dots, \numlayer}$ \citep{lederer2021activation}.

Neural networks,
such as those in~\eqref{eq:nn},
are typically fitted to data by minimizing the sum of loss function: $\min_{\matrices,\biases}\{\sum_{i=1}^{\numsample}\Loss_{\matrices,\biases}\bs{\outcomey_{i},\inputx_{i}}\}$.
The two standard loss functions for regression problems ($\numclass=1$) and classification problems ($\numclass \in \bc{2,3,\dots}$) are the  squared-error (SE) loss
\begin{equation}
\label{eq:se}
    \Loss_{\matrices, \biases}^{\se}\bs{\outcomey_{i},\inputx_{i}} 
    \eqv
    \prtb{\outcomey_{i} - \net_{\matrices,\biases}\bs{\samxi}}^{2}
\end{equation}
\noindent and the soft-max cross entropy (SCE) loss 
\begin{equation}
\label{eq:sce}
    \Loss_{\matrices, \biases}^{\sce}\bs{\outcomey_{i},\inputx_{i}} 
    \eqv
    -\prt{\outcomey_{i}}_{j}
    \log\bcbbbb{
    \frac{\exp\bcb{\prtb{\net_{\matrices,\biases}\bs{\samxi}}_{j}}}{\sum_{j=1}^{\numclass}\exp\bcb{\prtb{\net_{\matrices,\biases}\bs{\samxi}}_{j}}}
    } 
    ~~~~j \in \bc{1,\dots, \numclass},
\end{equation}
respectively.
It is well known that such loss functions efficient on benign data but sensitive to heavy-tailed data, corrupted samples, and so forth~\citep{huber:1964}.

We want to keep those loss functions' efficiency on benign samples, 
but, at the same time, avoid their failure in the presence of problematic samples.
We achieve this by a median-of-means approach ($\mom$) inspired by~\citep{Lecu2017RobustML}.
The details of the $\mom$ approach are  mathematically intricate,
but the general idea is simple.
We thus describe the general idea first and then formally define the estimator afterward.
The $\mom$ approach can roughly be formulated in terms  of three-step updates:
\begin{itemize}
\item[\textit{Step 1:}] Partition the data into blocks of samples.
\item[\textit{Step 2:}] On each block, calculate the empirical mean of the loss increment with respect to two separate sets of parameters $\prt{\matrices_{1},\biases_{1}}, \prt{\matrices_{2}, \biases_{2}} \in \prt{\matrixspace,\biaspace}$ and the standard loss function ($\Loss_{\matrices, \biases}^{\se}$ in regression and $\Loss_{\matrices, \biases}^{\sce}$ in classification).
\item[\textit{Step 3:}] Use the block that corresponds to the median of the empirical means in Step~2 to update the parameters.
\item[\textit{~~~}] \hspace{-7mm}Go back to Step~1 until convergence.
\end{itemize}

Let us now be more formal.
In the first step, we consider $\numblock \in \bc{1, \dots, \numsample}$ blocks  $\block_{1}, \dots, \block_{\numblock}\subset\{1,\dots,n\}$, that are an equipartition
of $\bc{1, \dots, \numsample}$, which means
the blocks have equal cardinalities 
$|\block_{1}|=\dots=|\block_{\numblock}|$, that
cover the entire index set $\cup_{k=1}^{\numblock}\block_{k} = \bc{1, \dots, \numsample}$. 
In practice, we set $|\block_{1}|=\dots=|\block_{\numblock-1}|=\lfloor\numsample/\numblock \rfloor$, where $\lfloor \numsample/\numblock \rfloor = \max\bc{\valuea \in \bc{1,2,\dots} : \valuea \leq \numsample/\numblock}$ and $|\block_{\numblock}| = \numsample - \sum_{i=1}^{\numblock-1} |\block_{i}|$, if $\numblock$ does not divide $\numsample$.

Given $\prt{\matrices_{1},\biases_{1}}, \prt{\matrices_{2}, \biases_{2}} \in \prt{\matrixspace,\biaspace}$, the quantities in Step~2 is defined by 
\begin{equation}
\label{eq:empericalMean}
    \prob_{\block_{k}} \bs{\prt{\matrices_{1},\biases_{1}}, \prt{\matrices_{2}, \biases_{2}}} 
    \eqv 
    \frac{1}{|\block_{k}|} \bsbbbb{\sum_{i \in \block_{k}} \Loss_{\matrices_{1}, \biases_{1}}\bs{\outcomey_{i},\samxi} - \Loss_{\matrices_{2}, \biases_{2}}\bs{\outcomey_{i},\samxi}} ~~~~~k \in \bc{1, \dots, \numblock},
\end{equation}
\noindent and we denote by $\quantile_{\block_{1}, \dots, \block_{\numblock}}^{\alpha}\bs{\prt{\matrices_{1},\biases_{1}}, \prt{\matrices_{2},\biases_{2}}}$ an $\alpha$-quantile of the set
\begin{equation*}
 \bcb{\prob_{\block_{1}}\bs{\prt{\matrices_{1},\biases_{1}}, \prt{\matrices_{2},\biases_{2}}}, \dots, \prob_{\block_{\numblock}} \bs{\prt{\matrices_{1},\biases_{1}}, \prt{\matrices_{2},\biases_{2}}}};
\end{equation*}
\noindent in particular, in Step~3, we compute the empirical median-of-means in Step~2 by defining 
\begin{align}
\label{eq:empiricalMedian}
\begin{split}
    &\mom_{\block_{1},\dots, \block_{\numblock}}\bs{\prt{\matrices_{1},\biases_{1}}, \prt{\matrices_{2},\biases_{2}}} 
    \eqv \\
    &\max_{k}\bigl\{
    \prob_{\block_{k}}\bs{\prt{\matrices_{1},\biases_{1}}, \prt{\matrices_{2},\biases_{2}}} :  \prob_{\block_{k}}\bs{\prt{\matrices_{1},\biases_{1}}, \prt{\matrices_{2},\biases_{2}}}  
    \leq
    \quantile_{\block_{1},\dots,\block_{\numblock}}^{1/2}\bs{\prt{\matrices_{1},\biases_{1}}, \prt{\matrices_{2},\biases_{2}}}
    \bigr\}.
\end{split}
\end{align}

Our estimator is then the solution of the min-max problem of the increment tests defined in the following.
\begin{definition}[\deepmom]
\label{def:estimator}
For $\numblock \in \bc{1, \dots, \numsample}$ and given blocks $\block_{1}, \dots, \block_{\numblock}$ described in the above, we define
\begin{equation*}
    \widehat{\weights}_{\mom}^{\block_{1},\dots, \block_{\numblock}} \eqv 
    \argmin_{\prt{\matrices_{1},\biases_{1}} \in \prt{\matrixspace,\biaspace}} 
    \sup_{\prt{\matrices_{2}\biases_{2}} \in \prt{\matrixspace,\biaspace}}
    \mom_{\block_{1},\dots, \block_{\numblock}}\bs{\prt{\matrices_{1},\biases_{1}}, \prt{\matrices_{2},\biases_{2}}}.
\end{equation*}
\end{definition}
\noindent
The rational is as follows: 
one the one hand,
using least-squares/cross-entropy on each block ensures efficient use of the ``good'' samples;
on the other hand,
using the median over the blocks removes the corruptions and, therefore, ensures robustness toward the ``bad'' samples.

In the case $\numblock = 1$, by definition, we have 
\begin{equation*}
\widehat{\weights}_{\mom}^{\block_{1}} \in 
\argmin\limits_{\prt{\matrices_{1},\biases_{1}} \in \prt{\matrixspace, \biaspace}} \Biggl\{
    \frac{1}{\numsample}
    \prtbbbb{
    \sum_{i=1}^{\numsample} 
        \Loss_{\matrices_{1},\biases_{1}}\bs{\outcomey_{i}, \samxi
        } - \\
        \sup_{\prt{\matrices_{2},\biases_{2}} \in \prt{\matrixspace,\biaspace}}
        \Loss_{\matrices_{2},\biases_{2}}\bs{\outcomey_{i}, \samxi
        }
    }
\Biggr\},
\end{equation*}
\noindent which implies that $\widehat{\weights}_{\mom}^{\block_{1}}$ is equivalent to the minimizer of $\sum_{i=1}^{\numsample}\Loss^{\se}_{\matrices_{1},\biases_{1}}\bs{\outcomey_{i}, \samxi}/\numsample$ or $\sum_{i=1}^{\numsample}\Loss^{\sce}_{\matrices_{1},\biases_{1}}\bs{\outcomey_{i}, \samxi}/\numsample$.
Hence, the $\deepmom$ estimator can also be seen as a generalization of the standard least-squares/cross-entropy approaches.

\section{Related literature}
\label{sec:RelateLiterature}

We now take a moment to highlight relationships with other approaches as well as differences to those approaches.
Since problematic samples are the rule rather than an exception in deep learning,
the sensitivity of the standard loss functions has sparked much research interest.
In regression settings, for example, \citep{barron2019general, belagiannis2015robust, jiang2018mentornet, wang2016studying, lederer2020risk} have replaced the squared-error loss by the absolute-deviation loss $\Loss_{\matrices, \biases}^{\abe}\bs{\outcomey_{i},\inputx_{i}}\eqv |\outcomey_{i}-\net_{\matrices,\biases}\bs{\samxi}|$, which generates estimators for the empirical median, or the Huber loss function~\citep{huber:1964, huber_ronchetti_2009, hampel_1986}
\begin{equation*}
    \Loss_{\matrices,\biases,\robust}^{\hue}\bs{\outcomey_{i},\inputx_{i}}
    \eqv
    \begin{cases}
      \frac{1}{2}\prtb{\outcomey_{i}-\net_{\matrices,\biases}\bs{\samxi}}^{2} 
      & |\outcomey_{i}-\net_{\matrices,\biases}\bs{\samxi}| \leq \robust\\
      \robust\prtb{|\outcomey_{i}-\net_{\matrices,\biases}\bs{\samxi}| - \frac{1}{2}\robust};
      & |\outcomey_{i}-\net_{\matrices,\biases}\bs{\samxi}| > \robust,
    \end{cases}
\end{equation*}
\noindent where $k\in(0,\infty)$ is a tuning parameter that determines the robustness of the function.
In classification settings, for example,
 \citep{goodfellow2015explaining,madry2019deep,wong2018provable} have added an $\ell_{1}$ penalty on the parameters.
Changing the loss function in those ways can make the estimators robust toward the problematic data,
but it also forfeits the efficiency of the standard loss functions in using the informative data.
In contrast, our approach offers robustness with respect to the ``bad'' samples but also efficient use of the ``good'' samples.

We, therefore, take a different route in this paper.
\citep{lugosi_mendelson_2019b,guillaume2017learning} 
have shown theoretically that median-of-means-based estimators can outperform standard least-squares estimators  when the data are heavy-tailed or corrupted.
Because these estimators are computationally infeasible,
\citep{lecue_lerasle_mathieu_2020, Lecu2017RobustML}~have replaced them with computationally tractable min-max versions and have shown that these estimators still achieve the sub-Gaussian deviation bounds of the earlier papers when the data are a generated by certain convex functions.
We transfer these ideas to the framework of deep learning, which, of course, is intrinsically non-convex and, therefore, mathematically and computationally more challenging.

Another related topic is adversarial attacks.
Adversarial attacks are intentional corruptions of the data with the goal of exploiting weaknesses of specific deep-learning pipelines~\citep{kurakin2016adversarial2,8376a3fbf7af40249bfa4109f33cea14,JMLR:v13:brueckner12a,Su_2019,athalye2018synthesizing}.
Hence, papers on adversarial attacks and our paper study data corruption.
However, the perspectives on corruptions are very different:
while the literature on adversarial attacks has a notion of a ``mean-spirited opponent,''
we have a much more general notion of ``good'' and ``bad'' samples.
The adversarial-attack perspective is much more common in deep learning,
but our view is much more common in the sciences more generally.
The different notions also lead to different methods:
methods in the context of adversarial attacks concern specific types of attacks and pipelines,
while our method can be seen as a way to render deep learning more robust in general.
A consequence of the different views is that adversarial attacks are designed for specific purposes,
while our approach can be seen as a general way to make deep learning more robust in general.
It would be misleading to include methods from the adversarial-attack literature in our numerical comparisons (they do not perform well simply because they are typically designed for very specific types of attacks and pipelines),
but one could use our method in adversarial-attack frameworks.
To avoid digression, we defer such studies to future work.

The following papers are related on a more general level: \citep{he2017adversarial} highlights that the combination of non-robust methods does not lead to a robust method.
\citep{carlini2017adversarial} shows that even the detection of problematic input is very difficult.
\citep{xu2010robustness} introduces a notion of algorithmic robustness to study  differences between training errors and expected errors. 
\citep{tramer2019adversarial} states that an ensemble of two robust methods, each of which is robust to a different form of perturbation, may be robust to neither.
\citep{tsipras2019robustness} demonstrates that there exists a trade-off between a model's standard accuracy and its robustness to adversarial perturbations.


\section{Algorithm and numerical analyses}
\label{sec:related}

In this section, we devise an algorithm for computing the \deepmom\ estimator of  Definition~\ref{def:estimator}.
We then corroborate in simulations that our estimator is both robust toward corruptions as well as efficient in using benign data.

\subsection{Algorithm}
\label{subsec:numerical}
It turns out that $\mom$ can be computed with standard optimization techniques.
In particular,
we can apply stochastic gradient steps.
The only minor challenge is that the estimator involves both a minimum and a supremum,
but this can be addressed by using two updates in each optimization step:
one update to make progress with respect to the minimum,
and one update to make progress with respect to the supremum.

To be more formal, we want to compute the estimator~$\widehat{\weights}_{\mom}^{\block_{1}, \dots, \block_{\numblock}}$ of Definition~\ref{def:estimator} for  given blocks $\block_{1}, \dots, \block_{\numblock}$ on data defined in Section~\ref{sec:estimator}.
This amounts to finding updates such that our objective function
\begin{equation*}
    \prtb{\prt{\matrices_{1},\biases_{1}}, \prt{\matrices_{2},\biases_{2}}} \mapsto\mom_{\block_{1},\dots, \block_{\numblock}}\bs{\prt{\matrices_{1},\biases_{1}}, \prt{\matrices_{2},\biases_{2}}}
\end{equation*}
\noindent descents in its first arguments $\prt{\matrices_{1},\biases_{1}}$ and ascents in its second arguments $\prt{\matrices_{2},\biases_{2}}$.
Hence, we are concerned with the gradients of 
\begin{equation}
\label{eq:first}
    \prt{\matrices_{1},\biases_{1}} \mapsto \mom_{\block_{1},\dots,\block_{\numblock}}\bs{\prt{\matrices_{1},\biases_{1}}, \prt{\matrices_{2}^{0},\biases_{2}^{0}}}
\end{equation}
\noindent and 
\begin{equation}
\label{eq:second}
    \prt{\matrices_{2},\biases_{2}} \mapsto \mom_{\block_{1},\dots, \block_{\numblock}}\bs{\prt{\matrices_{1}^{0}, \biases_{1}^{0}}, \prt{\matrices_{2},\biases_{2}}}
\end{equation}
\noindent for fixed $\prt{\matrices_{1}^{0},\biases_{1}^{0}}, \prt{\matrices_{2}^{0},\biases_{2}^{0}} \in \prt{\matrixspace, \biaspace}$. 

For simplicity, the gradient of an objective function $\func : \prt{\matrices,\biases} \mapsto \func\bs{\prt{\matrices,\biases}}$ with respect to $\prt{\matrices, \biases}$ at a point $\prt{\matrices^{0}, \biases^{0}}$ is denoted by $\nabla_{\prt{\matrices,\biases}}\func\bs{\prt{\matrices,\biases}}_{|\prt{\matrices,\biases}=\prt{\matrices^{0},\biases^{0}}}$, and the derivative of the activation functions $\actif^{j}$ are denoted by $\prt{\actif^{j}}'$.
(In line with the usual conventions,
the derivative of the ReLU function at zero is set to zero.)


Given $\prt{\matrices_{1}^{0}, \biases_{1}^{0}},\prt{\matrices_{2}^{0}, \biases_{2}^{0}} \in \prt{\matrixspace, \biaspace}$,
we can find---see the definition of the empirical median-of-means in~\eqref{eq:empiricalMedian}---indexes $k_{1}, k_{2} \in \bc{1, \dots, \numblock}$ (which depend on $\prt{\matrices_{1},\biases_{1}}, \prt{\matrices_{2},\biases_{2}} \in \prt{\matrixspace,\biaspace}$, respectively) such that $\prob_{\block_{k_{1}}} \bs{\prt{\matrices_{1},\biases_{1}}, \prt{\matrices_{2}^{0},\biases_{2}^{0}}}=\mom_{\block_{1},\dots,\block_{\numblock}}\bs{\prt{\matrices_{1},\biases_{1}}, \prt{\matrices_{2}^{0},\biases_{2}^{0}}}$ and $\prob_{\block_{k_{2}}} \bs{\prt{\matrices_{1}^{0}, \biases_{1}^{0}}, \prt{\matrices_{2},\biases_{2}}}=\mom_{\block_{1},\dots, \block_{\numblock}}\bs{\prt{\matrices_{1}^{0},\biases_{1}^{0}}, \prt{\matrices_{2},\biases_{2}}}$.

Therefore, given $\prt{\matrices_{11}^{0}, \biases_{11}^{0}}, \prt{\matrices_{22}^{0},\biases_{22}^{0}} \in \prt{\matrixspace, \biaspace}$, the gradients of~\eqref{eq:first} and~\eqref{eq:second} defined in Section~\ref{subsec:numerical} are
\begin{align*}
    &
    \nabla_{\matrices_{1},\biases_{1}} \mom_{\block_{1},\dots,\block_{\numblock}}
    \bs{\prt{\matrices_{1},\biases_{1}},\prt{\matrices_{2}^{0},\biases_{2}^{0}}}_{| \prt{\matrices_{1},\biases_{1}} =\prt{\matrices_{11}^{0}, \biases_{11}^{0}}} \\
    &= 
    \frac{1}{|\block_{k_{1}}|}
    \sum_{i \in \block_{k_{1}}}
    \nabla_{\matrices_{1},\biases_{1}}
    \prtb{
    \Loss_{\matrices_{1},\biases_{1}}\bs{\outcomey_{i},\samxi} - \Loss_{\matrices_{2}^{0},\biases_{2}^{0}}\bs{\outcomey_{i},\samxi}
    }_{|\prt{\matrices_{1},\biases_{1}}=\prt{\matrices_{11}^{0},\biases_{11}^{0}}} \\
    &=
    \frac{1}{|\block_{k_{1}}|}
    \sum_{i \in \block_{k_{1}}}
    \nabla_{\matrices_{1},\biases_{1}}
    \Loss_{\matrices_{1},\biases_{1}}\bs{\outcomey_{i},\samxi}_{|\prt{\matrices_{1},\biases_{1}}=\prt{\matrices_{11}^{0},\biases_{11}^{0}}}
\end{align*}
\noindent and 
\begin{align*}
    &
    \nabla_{\matrices_{2},\biases_{2}}
    \mom_{\block_{1},\dots,\block_{\numblock}}
    \bs{\prt{\matrices_{1}^{0},\biases_{1}^{0}},\prt{\matrices_{2},\biases_{2}}}_{| \prt{\matrices_{2},\biases_{2}} =\prt{\matrices_{22}^{0},\biases_{22}^{0}}} \\
    &= 
    \frac{1}{|\block_{k_{2}}|}
    \sum_{i \in \block_{k_{2}}}
    \nabla_{\matrices_{2},\biases_{2}}
    \prtb{
    \Loss_{\matrices_{1}^{0},\biases_{1}^{0}}\bs{\outcomey_{i},\samxi} - \Loss_{\matrices_{2},\biases_{2}}\bs{\outcomey_{i},\samxi}
    }_{|\prt{\matrices_{2},\biases_{2}}=\prt{\matrices_{22}^{0},\biases_{22}^{0}}} \\
    &= 
    -
    \frac{1}{|\block_{k_{2}}|}
    \sum_{i \in \block_{k_{2}}}
    \nabla_{\matrices_{2},\biases_{2}}
    \Loss_{\matrices_{2},\biases_{2}}\bs{\outcomey_{i},\samxi}_{|\prt{\matrices_{2},\biases_{2}}=\prt{\matrices_{22}^{0},\biases_{22}^{0}}}.
\end{align*}

The individual gradients $\nabla_{\matrices,\biases}\Loss_{\matrices, \biases}\bs{\outcomey_{i},\samxi}_{|\prt{\matrices,\biases}=\prt{\matrices^{0}, \biases^{0}}}$ can then be computed by back propagation~\citep{rumelhart_hinton_williams_1986}.

The above computations are then the basis for our computation of $\widehat{\weights}_{\mom}^{\block_{1},\dots,\block_{\numblock}}$ in Algorithm~\ref{alg:mom}.
In that algorithm, we set 
\begin{equation*}
    \norm{\prt{\matrices_{1},\biases_{1}}-\prt{\matrices_{2},\biases_{2}}}_{2} 
    \eqv 
    \sqrt{
    \sum_{j=0}^{\numlayer} 
    \sum_{v=1}^{\numparameter^{j+1}}
    \sum_{w=1}^{\numparameter^{j}}
    \prtb{[\matri_{1}^{j}-\matri_{2}^{j}]_{v,w}}^{2}
    +
    \sum_{j=0}^{\numlayer} 
    \sum_{v=1}^{\numparameter^{j+1}}
    \prtb{[\bias_{1}^{j}-\bias_{2}^{j}]_{v}}^{2}
    }
\end{equation*}
for $\prt{\matrices_{1},\biases_{1}}, \prt{\matrices_{2},\biases_{2}} \in \prt{\matrixspace,\biaspace}$.

\begin{algorithm}
    \footnotesize
	\caption{stochastic gradient-based algorithm for \deepmom}
	\label{alg:mom}
	\begin{algorithmic}
		\STATE {\bfseries Input:} data $\prt{\outcomey_1,\boldsymbol{x}_1},\dots,\prt{\outcomey_{\numsample}, \boldsymbol{x}_{\numsample}}$,
		number of blocks $\numblock$,
		initial parameters $\prt{\matrices_{1}^{0},\biases_{1}^{0}}, \prt{\matrices_{2}^{0},\biases_{2}^{0}}$, \\ iteration number $\imax$, stopping criterion $\stopping$, batch size $\batch$, and learning rate $\rate$. \\ 
		\STATE {\bfseries Output:} $\widehat{\weights}_{\mom}^{\block_{1},\dots,\block_{\numblock}}$ of Definition~\ref{def:estimator}. \\
		\WHILE{$i \leq \imax$}
		    \STATE {\bfseries 1.} Randomly select a batch of $\batch$ data points. \\
		    \STATE {\bfseries 2.} Generate blocks $\block_{1},\dots, \block_{\numblock}$ for the selected data. \\
    		\STATE {\bfseries 3.} Update gradients for the first arguments : \\
    		~~~~~$\prt{\matrices_{1}^{i+1},\biases_{1}^{i+1}} 
    		\eqv 
    		\prt{\matrices_{1}^{i},\biases_{1}^{i}} - \rate \nabla_{\matrices_{1},\biases_{1}} \mom_{\block_{1},\dots,\block_{\numblock}}
    		\bs{\prt{\matrices_{1},\biases_{1}},\prt{\matrices_{2}^{i},\biases_{2}^{i}}}_{| \prt{\matrices_{1},\biases_{1}} =\prt{\matrices_{1}^{i},\biases_{1}^{i}}}$ \\
    		
    		\STATE {\bfseries 4.} First stopping criterion:
    		~~~~~\IF{$\norm{\prt{\matrices_{1}^{i+1},\biases_{1}^{i+1}}-\prt{\matrices_{1}^{i},\biases_{1}^{i}}}_{2} \leq \stopping$}
    		    \STATE  {\bfseries break}
    		\ENDIF
    		\\
    		\STATE {\bfseries 5.} Update gradients for the second arguments : \\
    		~~~~~$\prt{\matrices_{2}^{i+1},\biases_{2}^{i+1}} 
    		\eqv
    		\prt{\matrices_{2}^{i},\biases_{2}^{i}}-\rate
    		\nabla_{\matrices_{2},\biases_{2}}\mom_{\block_{1},\dots,\block_{\numblock}}\bs{\prt{\matrices_{1}^{i+1},\biases_{1}^{i+1}},\prt{\matrices_{2},\biases_{2}}}_{| \prt{\matrices_{2},\biases_{2}}=\prt{\matrices_{2}^{i},\biases_{2}^{i}}}$ \\
    		
    		\STATE {\bfseries 6.} Second stopping criterion:
    		~~~~~\IF{$\norm{\prt{\matrices_{2}^{i+1},\biases_{2}^{i+1}}-\prt{\matrices_{2}^{i}}_{2},\biases_{2}^{i}}_{2} \leq \stopping$}
    		    \STATE  {\bfseries break}
    		\ENDIF
		\ENDWHILE
	\end{algorithmic}
\end{algorithm}

Throughout the paper, the batch size is $\batch=0.15 \numsample$, maximum number of iteration $\imax=20\,000$, and the stopping criterion $\stopping=10^{-2}$.

\begin{figure}[htb!]
  \center
  \includegraphics[width=0.5\columnwidth]{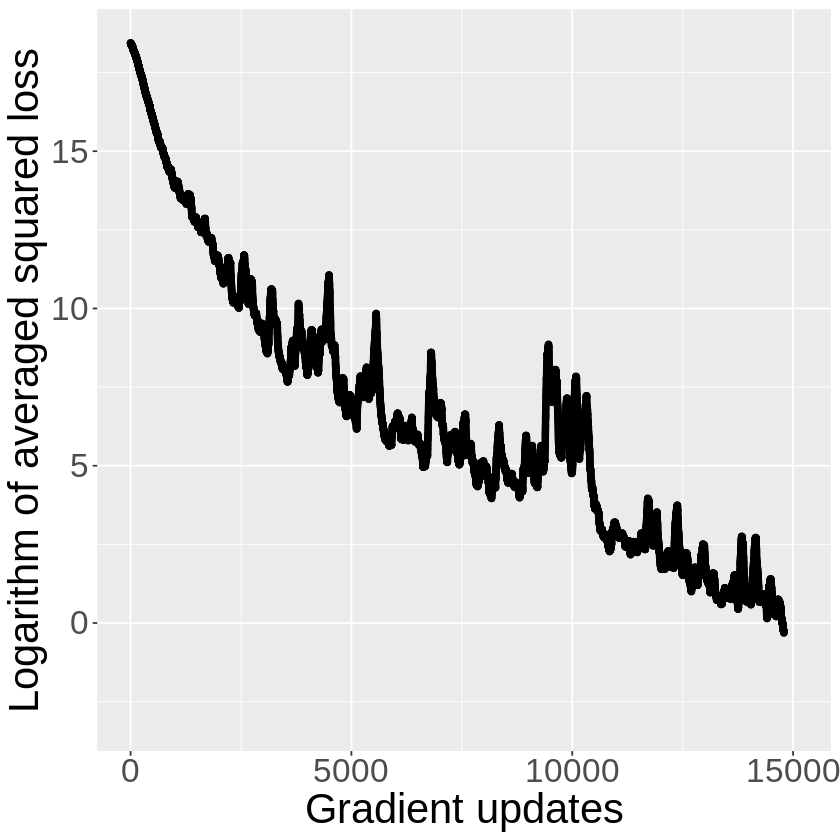}
\caption{the logarithm training loss of $\deepmom$ (as in Algorithm~\ref{alg:mom}) with $\prt{\numsample, \numparameter, \numlayer, \numneuron}=\prt{2000, 50, 7, 30}$, $|\info|/\numsample=0.75$, and $\numblock=5$ as a function of the gradient updates
}
\label{fig:loss}
\end{figure}

Algorithm~\ref{alg:mom} provides a stochastic approximation method for standard gradient-descent optimization of the empirical-median-of-means function formulated in Display~\eqref{eq:empiricalMedian}.
An illustration for the convergence of Algorithm~\ref{alg:mom} is shown in Figure~\ref{fig:loss} and the mathematical result on the convergence of the algorithm is as follows..

\begin{proposition}[Convergence]
\label{prop:convergence}

If there exist parameters $\prt{\matrices, \biases} \in \prt{\matrixspace,\biaspace}$ that make the objective function given in~\eqref{eq:empiricalMedian} equal to zero, Algorithm~\ref{alg:mom} is guaranteed to converge.
\end{proposition}
\begin{proof}[Proof of Proposition~\ref{prop:convergence}]
\label{prf:prop1}
The proof follows readily from \citep{liu2019accelerating, sun2019optimization}.
\end{proof}

The assumption of parameters for a perfect fit can be relaxed,
but in this paper,
we are interested less in the mathematical details of our specific algorithm but rather in showing that the $\mom$ approach works in the first place.
Our numerical studies in the following serve this purpose.

\subsection{Numerical analyses for regression data}
\label{subsec:simulation1}

We now consider the regression data and show that our $\mom$ approach can indeed outmatch other approaches,
such as vanilla least-squares estimation, absolute deviation, and Huber estimation.

\paragraph{General setup} 
We consider a uniform width $\numneuron\eqv\numparameter^{1}=\dots=\numparameter^{\numlayer}$.
The elements of the input vectors $\boldsymbol{x}_{1}, \dots, \boldsymbol{x}_{\numsample}$ are i.i.d.~sampled from a standard Gaussian distribution
and then normalized such that
$\sum_{i=1}^{\numsample}((\boldsymbol{x}_{i})_j)^2=1$  for all $j\in\{1,\dots,\numparameter\}$. 
The elements of the true weight matrices $\matrices^* = \prt{\matri^{0}, \dots, \matri^{\numlayer}}$ and true bias parameters $\biases^* = \prt{\bias^{0}, \dots, \bias^{\numlayer}}$ are i.i.d.~sampled from a uniform distribution on $[-1,1]$.
The observations of the stochastic noise variables  $\err_1,\dots,\err_{\numsample}$ are i.i.d.~sampled from a centered Gaussian distribution such that the empirical signal to noise ratio equals $10$.

\paragraph{Data corruptions}
We corrupt the data in three different ways.
Recall that~$\info$ and~$\outlier$ denote the sets of informative samples and corrupted samples (outliers), respectively.

\emph{Corrupted outputs (outliers):}
The noise variables $\err_{i}$ for outliers $i\in\outlier$ are replaced by i.i.d.~samples from a uniform distribution on  $[3\max_{i}|\net_{\matrices^{*},\biases^{*}}\prt{\samxi}|, 5\max_{i}|\net_{\matrices^{*},\biases^{*}}\prt{\samxi}|]$.
This means that the corresponding outputs~$\outcomey_{i}$ are subject to heavy yet bounded corruptions.

\emph{Corrupted outputs (everywhere):}
All noise variables are replaced by i.i.d.~samples from a Student's t-distribution with $df$~degrees of freedom.
This means that all outputs~$\outcomey_{i}$ are subject to unbounded corruptions.

\emph{Corrupted inputs:}
The elements of the input vectors $\samxi$ for outliers $i\in\outlier$ receive (after computing~$\outcomey_{i}$) an additional perturbation that is i.i.d.~sampled from a standard Gaussian distribution.
This means that the analyst gets to see corrupted versions of the input.

\begin{table}[ht]
    \center
	\footnotesize
	\setlength\tabcolsep{3.5pt}
	\begin{tabular}{c c c c c}
	
		\toprule
		\midrule

		\multicolumn{5}{c}{$\prt{\numsample, \numparameter, \numlayer, \numneuron}=\prt{1000, 50, 5, 50}$} 
		\\ \cline{1-5}
		\multicolumn{5}{c}{Corrupted outputs (uniform distribution)} 
		\\ \cline{1-5}
		{} &
		\multicolumn{4}{c}{Scaled mean of prediction errors} 
		\\ \cline{2-5}

		\multicolumn{1}{c}{$|\info|/\numsample$} &
		\multicolumn{1}{c}{$\Mmin$} & 
		\multicolumn{1}{c}{$\abe$} & 
		\multicolumn{1}{c}{$\Hmin$} & 
		\multicolumn{1}{c}{$\se$} \\
		
		\hline
	 
	    $100\,\%$ & 1.00 & \hphantom{0}1.58 & \hphantom{0}1.29 & \hphantom{0}\hphantom{0}1.60  \\
	    $95\,\%$ & 1.22 & 12.09 & \hphantom{0}7.36 & \hphantom{0}17.42  \\
	    $85\,\%$ & 1.48 & 26.81 & 33.69 & \hphantom{0}72.21  \\
	    $75\,\%$ & 2.47 & 80.34 & 68.76 & 121.58  \\
		\midrule
		\midrule
		
		\multicolumn{5}{c}{Corrupted outputs (t-distribution)} 
		\\ \cline{1-5}
		{} &
		\multicolumn{4}{c}{Scaled mean of prediction errors} 
		\\ \cline{2-5}
		
		df &
		\multicolumn{1}{c}{$\Mmin$} & 
		\multicolumn{1}{c}{$\abe$} & 
		\multicolumn{1}{c}{$\Hmin$} & 
		\multicolumn{1}{c}{$\se$} \\
		
		\hline
	 
	    10 & 1.16 & 1.40 & 1.34 & 1.93  \\
	    1 & 1.11 & 1.38 & 1.32 & 1.83  \\
	    
	    \midrule
		\midrule
		
		\multicolumn{5}{c}{Corrupted inputs} 
		\\ \cline{1-5}
		{} &
		\multicolumn{4}{c}{Scaled mean of prediction errors} 
		\\ \cline{2-5}
		
		\multicolumn{1}{c}{$|\info|/\numsample$} &
		\multicolumn{1}{c}{$\Mmin$} & 
		\multicolumn{1}{c}{$\abe$} & 
		\multicolumn{1}{c}{$\Hmin$} & 
		\multicolumn{1}{c}{$\se$} \\
		
		\hline
	 
	    $95\,\%$ & 1.26 & 1.46 & 1.77 & 1.61  \\
	    $85\,\%$ & 1.27 & 1.38 & 1.38 & 1.80  \\
	    $75\,\%$ & 1.37 & 1.66 & 1.90 & 1.75  \\
	    
		\midrule
		\bottomrule
	\end{tabular}
	\captionof{table}{$\Mmin$ outperforms $\widehat{\weight}_{\se}$, $\widehat{\weight}_{\abe}$, and $\Hmin$ in all settings}
\label{table:simc11}
\end{table}

\begin{table}[ht]
    \center
	\footnotesize
	\setlength\tabcolsep{3.5pt}
	\begin{tabular}{c c c c c}
		\toprule
		\midrule
		
		\multicolumn{5}{c}{$\prt{\numsample, \numparameter, \numlayer, \numneuron}=\prt{1600, 50, 10, 20}$} 
		\\ \cline{1-5}
		\multicolumn{5}{c}{Corrupted outputs (uniform distribution)} 
		\\ \cline{1-5}
		{} &
		\multicolumn{4}{c}{Scaled mean of prediction errors} 
		\\ \cline{2-5}
		
		\multicolumn{1}{c}{$|\info|/\numsample$} &
		\multicolumn{1}{c}{$\Mmin$} & 
		\multicolumn{1}{c}{$\abe$} & 
		\multicolumn{1}{c}{$\Hmin$} & 
		\multicolumn{1}{c}{$\se$} \\
		
		\hline
	 
	    $100\,\%$ & 1.00 & \hphantom{0}\hphantom{0}2.04 & \hphantom{0}\hphantom{0}1.71 & \hphantom{0}\hphantom{0}3.62  \\
	    $95\,\%$ & 1.58 & \hphantom{0}19.85 & \hphantom{0}10.66 & \hphantom{0}31.66  \\
	    $85\,\%$ & 1.73 & \hphantom{0}63.69 & \hphantom{0}70.71 & 116.85  \\
	    $75\,\%$ & 2.06 & 159.17 & 133.71 & 229.14  \\
	    
	    \midrule
		\midrule
		
		\multicolumn{5}{c}{Corrupted outputs (t-distribution)} 
		\\ \cline{1-5}
		{} &
		\multicolumn{4}{c}{Scaled mean of prediction errors} 
		\\ \cline{2-5}
		
		df &
		\multicolumn{1}{c}{$\Mmin$} & 
		\multicolumn{1}{c}{$\abe$} & 
		\multicolumn{1}{c}{$\Hmin$} & 
		\multicolumn{1}{c}{$\se$} \\
		
		\hline
	    
	    10 & 1.05 & 1.72 & 1.75 & 2.83  \\
	    1 & 1.99 & 2.18 & 1.81 & 3.93  \\
	    
		\midrule
		\midrule
		
		\multicolumn{5}{c}{Corrupted inputs} 
		\\ \cline{1-5}
		{} &
		\multicolumn{4}{c}{Scaled mean of prediction errors} 
		\\ \cline{2-5}
		
		\multicolumn{1}{c}{$|\info|/\numsample$} &
		\multicolumn{1}{c}{$\Mmin$} & 
		\multicolumn{1}{c}{$\abe$} & 
		\multicolumn{1}{c}{$\Hmin$} & 
		\multicolumn{1}{c}{$\se$} \\
		
		\hline
	    
	    $95\,\%$ & 1.63 & 2.37 & 1.97 & 2.70  \\
	    $85\,\%$ & 1.75 & 2.08 & 2.53 & 3.21  \\
	    $75\,\%$ & 1.86 & 2.22 & 2.63 & 3.78  \\
	    
		\midrule
		\bottomrule
	\end{tabular}
	\captionof{table}{$\Mmin$ outperforms $\widehat{\weight}_{\se}$, $\widehat{\weight}_{\abe}$, and $\Hmin$ in  all settings}
\label{table:simc12}
\end{table}

\paragraph{Error quantification}
$20$ data sets are generated as described above.
The first half of the samples in each data set are assigned to training
and the remaining half of the samples to testing.
For each estimator~$\widehat{\weights}$,
the average of the generalization error
\begin{equation*}
    2/\numsample\sum_{i=\frac{\numsample}{2} + 1}^{\numsample}\prt{\net_{\matrices^*, \biases^*}\bs{\samxi}-\net_{\widehat{\weights}}\bs{\samxi}}^2
\end{equation*}
\noindent is computed and re-scaled with respect to the $\mom$ approach with $100\%$~informative data.

The contenders are $\mom$, absolute error, Huber, and the squared error loss functions.
For convenience, we denote $\widehat{\weights}_{\se}$, $\widehat{\weights}_{\abe}$, and $\widehat{\weights}_{\hue}^{\robust}$ as the estimators obtained by minimizing  $2\sum_{i=1}^{\numsample}\Loss_{\matrices,\biases}^{\se}\bs{\outcomey_{i},\samxi}/\numsample$, $2\sum_{i=1}^{\numsample}\Loss_{\matrices,\biases}^{\abe}\bs{\outcomey_{i},\samxi}/\numsample$, and
$2\sum_{i=1}^{\numsample}\Loss_{\matrices,\biases,\robust}^{\hue}\bs{\outcomey_{i},\samxi}/\numsample$ on training data, respectively.

Besides, we consider a sequence of $\mom$ estimators $\widehat{\weights}_{\mom}^{\block_{1},\dots,\block_{\numblock}}$, where $\numblock \in \bc{1,21,\dots,121}$, and we define $\Mmin$ as
\begin{equation*}
    \Mmin \eqv 
    \min_{\numblock}\bcbb{
    \frac{2}{\numsample}\sum_{i=\frac{2}{\numsample}+1}^{\numsample}
    \prtb{\net_{\matrices^*, \biases^*}\bs{\samxi} - \net_{\widehat{\weights}_{\mom}^{\block_{1},\dots,\block_{\numblock}}}\bs{\samxi}}^{2}
    }.
\end{equation*}

We further consider a sequence of Huber estimators $\widehat{\weights}_{\hue}^{\robust_{q}}$, where $k_{q}$ are the q-th quantile of $\bc{|\outcomey_{i}|}_{i \in \bc{1, \dots, \numsample}}$ with $q \in \bc{75, 80, 85, 90, 95, 100}$, and we define $\Hmin$ as 
\begin{equation*}
    \Hmin \eqv 
    \min_{q}\bcbb{
    \frac{2}{\numsample}\sum_{\frac{2}{\numsample}+1}^{\numsample}
    \prtb{\net_{\matrices^*, \biases^*}\bs{\samxi} - \net_{\widehat{\weights}_{\hue}^{k_{q}}}\bs{\samxi}}^{2}
    }.
\end{equation*}

\paragraph{Results and conclusions} 
The results for different settings are summarized in Tables~\ref{table:simc11}--\ref{table:simc13}.
First, we observe that $\mom$, least-squares, and Huber estimators behave very similarly in the uncorrupted case ($|\info| / \numsample=100\%$) and for mildly corrupted outputs ($df=10$).
But once the corruptions are more substantial, 
our $\mom$ approach clearly outperforms the other approaches.
In general,
we conclude that $\mom$ is efficient on benign data and robust on problematic data.

\begin{table}[!hb]
    \center
	\setlength\tabcolsep{3.5pt}
	\begin{tabular}{c c c c c}
	
		\toprule
		\midrule

        \multicolumn{5}{c}{$\prt{\numsample, \numparameter, \numlayer, \numneuron}=\prt{2000, 50, 7, 30}$} 
		\\ \cline{1-5}
		\multicolumn{5}{c}{Corrupted outputs (uniform distribution)} 
		\\ \cline{1-5}
		{} &
		\multicolumn{4}{c}{Scaled mean of prediction errors} 
		\\ \cline{2-5}
		
		\multicolumn{1}{c}{$|\info|/\numsample$} &
		\multicolumn{1}{c}{$\Mmin$} & 
		\multicolumn{1}{c}{$\abe$} & 
		\multicolumn{1}{c}{$\Hmin$} & 
		\multicolumn{1}{c}{$\se$} \\
		
		\hline
	 
	    $100\,\%$ & 1.00 & \hphantom{0}1.74 & \hphantom{0}1.53 & \hphantom{0}2.05  \\
	    $95\,\%$ & 1.05 & 14.34 & 10.84 & 14.45  \\
	    $85\,\%$ & 1.39 & 44.80 & 44.29 & 44.58  \\
	    $75\,\%$ & 1.95 & 78.19 & 95.02 & 82.56  \\
		
		\midrule
		\midrule
		
		\multicolumn{5}{c}{Corrupted outputs (t-distribution)} 
		\\ \cline{1-5}
		{} &
		\multicolumn{4}{c}{Scaled mean of prediction errors} 
		\\ \cline{2-5}
		
		df &
		\multicolumn{1}{c}{$\Mmin$} & 
		\multicolumn{1}{c}{$\abe$} & 
		\multicolumn{1}{c}{$\Hmin$} & 
		\multicolumn{1}{c}{$\se$} \\
		
		\hline
	    
	    10 & 0.97 & 1.58 & 1.70 & 2.44  \\
	    1 & 0.99 & 1.68 & 1.78 & 2.05  \\
		
		\midrule
		\midrule
		
		\multicolumn{5}{c}{Corrupted inputs} 
		\\ \cline{1-5}
		{} &
		\multicolumn{4}{c}{Scaled mean of prediction errors} 
		\\ \cline{2-5}
		
		\multicolumn{1}{c}{$|\info|/\numsample$} &
		\multicolumn{1}{c}{$\Mmin$} & 
		\multicolumn{1}{c}{$\abe$} & 
		\multicolumn{1}{c}{$\Hmin$} & 
		\multicolumn{1}{c}{$\se$} \\
		
		\hline
	    
	    $95\,\%$ & 1.01 & 1.60 & 1.76 & 1.99  \\
	    $85\,\%$ & 1.07 & 1.66 & 1.58 & 1.87  \\
	    $75\,\%$ & 1.13 & 1.58 & 1.38 & 2.13  \\
	    
		\midrule
		\bottomrule
	\end{tabular}
	\captionof{table}{$\Mmin$ outperforms $\widehat{\weight}_{\se}$, $\widehat{\weight}_{\abe}$, and $\Hmin$ in all settings}
\label{table:simc13}
\end{table}


\subsection{Numerical analyses for classification data}
\label{subsec:simulation2}

We now demonstrate that the $\deepmom$ estimator in Definition~\ref{def:estimator} also outperforms  soft-max cross-entropy estimation in multiclass classification.

\begin{figure}[htb!]
  \center
  \includegraphics[width=0.4\columnwidth]{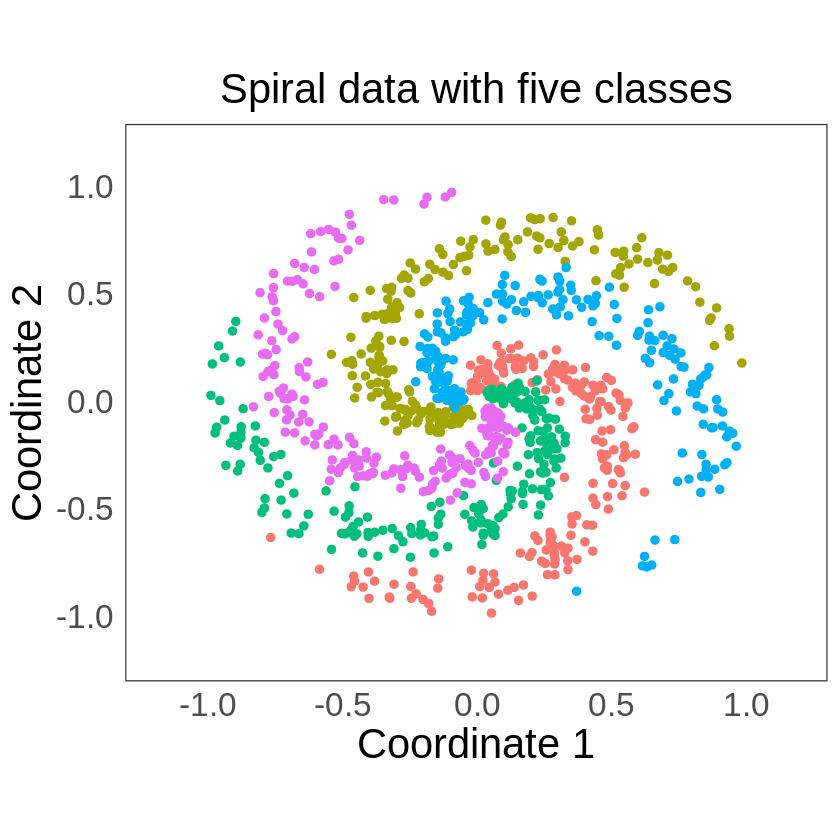}
\caption{the two-dimensional spiral data set with five classes}
\label{fig:spiral}
\end{figure}

\paragraph{General setup} 
We consider a spiral data set inspired by~\citep{Guan2019DCGAD, Amiri2015OnTE, Helfmann2018OnHS}.
The dimensionality is $\prt{\numsample, \numparameter} = \prt{1000,2}$,
and index set is partitioned into five classes $\class_{1}, \dots, \class_{5}$ with identical cardinalities: $|\class_{1}|=\cdots=|\class_{5}|=200$. 
Denoting the Hadamard product between two vectors by $\circ$ and the normal distribution with mean $\mu$ and standard deviation $\sigma$ by $\normal\prt{\mu,\sigma^{2}}$,
we construct the data as follows:
for each class $j \in \bc{1,\dots, 5}$, we set 
\begin{equation*}
    \prt{\outcomey_{i}}_{k}
    \eqv
    \begin{cases}
      1
      & k=j\\
      0
      & k \neq j 
    \end{cases}~~~~~ k \in \bc{1, \dots, 5},~~ i \in \class_{j},
\end{equation*}
\noindent $\prt{\prt{\samxi}_{1}}_{i \in \class_{j}} \eqv \radius \circ \sin\prt{\target}$, and $\prt{\prt{\samxi}_{2}}_{i \in \class_{j}} \eqv \radius \circ \cos\prt{\target}$, 
where the elements of $\radius$ and $\target$ are given by $\radius_{m} \eqv 0.05 + \prt{1-0.05}\prt{m-1}/200$ and $\prt{\target}_{m} \eqv \prt{j-1}3.7 + 3.7\prt{m-1}/200 + u_{m}$ for $u_{m} \sim \normal[0,0.25]$ and $m \in \bc{1,\dots, 200}$.
Each element of each input vector $\boldsymbol{x}_{1}, \dots, \boldsymbol{x}_{\numsample}$ is then divided by the maximum among the elements of that vector,  
so that $\max_{i,j}\prt{\samxi}_{j} = 1$ for all $i \in \bc{1,\dots, \numsample/2}$ and $j\in\bc{1,2}$. 
The data are visualized in Figure~\ref{fig:spiral}. 

To fit these data, 
we consider a two-layer ReLU network with 
uniform width $\numneuron\eqv\numparameter^{1}=\numparameter^{2}=150$.

\begin{table}[htb]
\centering
	\begin{tabular}{c c c}
	
		\toprule
		\midrule
		
		\multicolumn{3}{c}{Corrupted outputs (shuffled labels)} 
		\\ \cline{1-3}
		{} &
		\multicolumn{2}{c}{Prediction accuracies} 
		\\ \cline{2-3}

		\multicolumn{1}{c}{$|\info|/\numsample$} &
		\multicolumn{1}{c}{$\Mmin$} & 
		\multicolumn{1}{c}{$\sce$}  \\
		
		\hline
	 
	    $100\,\%$ & 95.6\,\% & 95.6\,\%   \\
	    $95\,\%$ & 95.2\,\% & 94.2\,\%   \\
	    $85\,\%$ & 88.0\,\% & 79.6\,\%   \\
	    $75\,\%$ & 85.6\,\% & 73.2\,\%   \\
		\midrule
		\midrule
		
		\multicolumn{3}{c}{Corrupted inputs} 
		\\ \cline{1-3}
		{} &
		\multicolumn{2}{c}{Prediction accuracies} 
		\\ \cline{2-3}
		
		\multicolumn{1}{c}{$|\info|/\numsample$} &
		\multicolumn{1}{c}{$\Mmin$} & 
		\multicolumn{1}{c}{$\sce$} \\
		
		\hline
	 
	    $95\,\%$ & 95.6\,\% & 95.6\,\%   \\
	    $85\,\%$ & 95.6\,\% & 95.6\,\%   \\
	    $75\,\%$ & 93.8\,\% & 90.2\,\%   \\
	    
		\midrule
		\bottomrule
	\end{tabular}
	\caption{table}{$\Mmin$ rivals or outperforms $\widehat{\weight}_{\sce}$ in all settings}
\label{table:simclass}
\end{table}

\paragraph{Data corruptions}
We corrupt the spiral data in two ways:

\emph{Corrupted labels:}
The labels $\outcomey_{i}$ for $i \in \outlier$ are changed to other class labels.

\emph{Corrupted inputs:}
The elements of the input vectors $\samxi$ for outliers $i\in\outlier$ are subjected to additional perturbations as described in Section~\ref{subsec:simulation1}.

\paragraph{Error quantification}
The first half of the samples will be used for training, while the other half will be used for testing.
For the $\sce$ estimator~$\widehat{\weights}_{\sce}$, which is computed by minimizing the mean of $\Loss_{\matrices, \biases}^{\sce}\bs{\outcomey_{i}, \samxi}$ on the training data,
the generalization accuracy $2 \sum_{i={\numsample}/{2} +  1}^{\numsample}\mathbbm{1}_{\prt{\outcomey_{i}=\net_{\widehat{\weights}_{\sce}}\bs{\samxi}}}/\numsample$ is calculated, where $\mathbbm{1} : \R^{\numclass} \mapsto \R$ is the indicator function defined by
\begin{equation*}
    \mathbbm{1}_{\prt{\outcomey_{i}=\net_{\widehat{\weights}_{\sce}}\bs{\samxi}}}
    \eqv
    \begin{cases}
      1
      & \outcomey_{i}=\net_{\widehat{\weights}_{\sce}}\bs{\samxi}\\
      0
      & \outcomey_{i}\neq\net_{\widehat{\weights}_{\sce}}\bs{\samxi} 
    \end{cases}~~~~~ i \in \bcbb{\frac{\numsample}{2},\dots, \numsample}.
\end{equation*}

We define $\Mmin$ as in Section~\ref{subsec:simulation1},
where the number of blocks~$\numblock$ can range in $\bc{1,3,\dots,11}$.

\paragraph{Results and conclusions} 
Table~\ref{table:simclass} summarizes the results for various settings.
To begin, we see that $\deepmom$ and $\sce$ estimators behave very similarly in the case of uncorrupted and slightly corrupted labels ($|\info| / \numsample=100\%, 95\%$) and slightly corrupted inputs ($|\info| / \numsample=95\%,85\%$).
But when the corruptions are more serious, our $\deepmom$ estimator clearly outperforms the other approaches.
In general,
we conclude that $\mom$ is efficient on benign data and robust on problematic data.

\section{Discussion}
\label{sec:discussion}
Our new approach to training the parameters of neural networks is robust against corrupted samples and yet leverages informative samples efficiently.
We have confirmed these properties numerically in Sections~\ref{sec:real}, \ref{subsec:simulation1}, and~\ref{subsec:simulation2}.
The approach can, therefore, be used as a general substitute for basic least-squares-type or cross-entropy-type approaches.

We have restricted ourselves to feed-forward neural networks with ReLU activation,
but there are no obstacles for applying our approach more generally,
for example, to convolutional networks or other activation functions.
However, to keep the paper clear and concise,
we defer a detailed analysis of $\mom$ in other deep-learning frameworks to future work.

Similarly, we model corruption by uniform or heavy-tailed random perturbations of the inputs or outputs or by randomly swapping labels,
but, of course,
one can conceive a plethora of different ways to corrupt data.

In sum, given modern data's limitations and our approach's ability to make efficient use of such data,
we believe that our method can have a substantial impact on deep-learning practice,
and that our paper can spark further interest in robust deep learning.

\subsection*{Acknowledgements}
We thank Guillaume Lecu\'e, Timoth\'e Mathieu, Mahsa Taheri, and Fang Xie for their insightful inputs an suggestions.

\bibliographystyle{Perfect}
\bibliography{Bibliography-MM-MC}



\end{document}